%% file: 0_main.tex
\documentclass[10pt,twocolumn,letterpaper]{article}

\usepackage[pagenumbers]{cvpr} % To force page numbers, e.g. for an arXiv version

\usepackage{graphicx}
\usepackage{amsmath}
\usepackage{amssymb}
\usepackage{booktabs}
\usepackage{enumitem}

\usepackage[pagebackref,breaklinks,colorlinks]{hyperref}

\usepackage[capitalize]{cleveref}
\crefname{section}{Sec.}{Secs.}
\Crefname{section}{Section}{Sections}
\Crefname{table}{Table}{Tables}
\crefname{table}{Tab.}{Tabs.}

\def\cvprPaperID{10286} % *** Enter the CVPR Paper ID here
\def\confName{CVPR}
\def\confYear{2023}

\begin{document}

\title{From Coarse to Fine-grained Concept based Discrimination for Phrase Detection}

\author{Maan Qraitem, Bryan A. Plummer\\
Boston University\\
{\{mqraitem, bplum\}@bu.edu}
}
\maketitle
\input{1_abstract}

\input{2_introduction}

\input{3_related_work}

\input{4_method}

\input{5_experiments}

%\input{7_broader_impacts}
\input{6_conclusion}
%\input{7_ethics_rep}

%%%%%%%%% REFERENCES
{\small
\bibliographystyle{ieee_fullname}
\bibliography{egbib}
}

\end{document}

%% file: 1_abstract.tex
\begin{abstract}
    Phrase detection requires methods to identify if a phrase is relevant to an image and localize it, if applicable. A key challenge for training more discriminative detection models is sampling negatives. Sampling techniques from prior work focus primarily on hard, often noisy, negatives disregarding the broader distribution of negative samples. Our proposed CFCD-Net addresses this through two novels methods. First, we generate groups of semantically similar words we call concepts (\eg, \{dog, cat, horse\} and \ \{car, truck, SUV\}), and then train our CFCD-Net to discriminate between a region of interest and its unrelated concepts. Second, for phrases containing fine-grained mutually-exclusive words (\eg, colors), we force the model to select only one applicable phrase for each region using our novel fine-grained module (FGM). We evaluate our approach on Flickr30K Entities and RefCOCO+, where we improve mAP over the state-of-the-art by 1.5-2 points. When considering only the phrases affected by our FGM module, we improve by 3-4 points on both datasets.
\end{abstract}

%% file: 2_introduction.tex
\section{Introduction}

\begin{figure}[t!]
\centering
\includegraphics[width=0.9\linewidth]{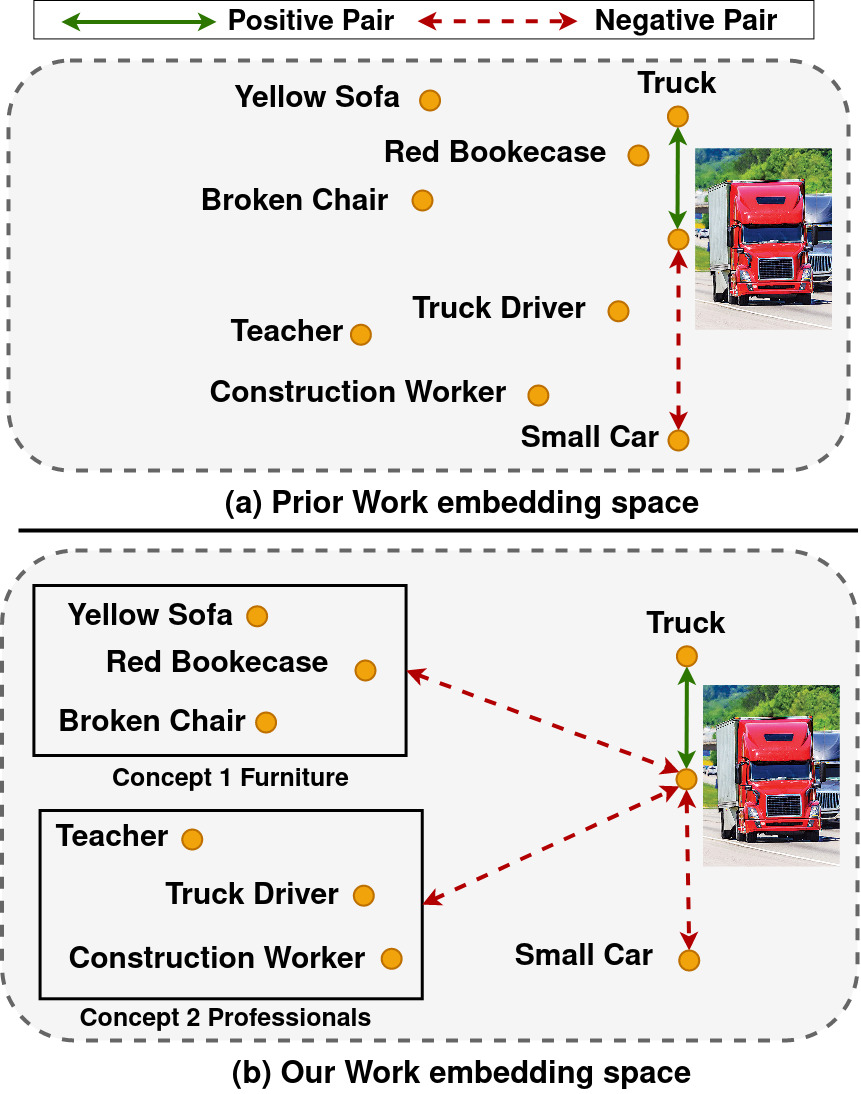}
\vspace{-2mm}
\caption{\textbf{Comparing embedding spaces}. (a) prior work used hard-negative mining to make their model more discriminative (\eg,~\cite{hinami-satoh-2018-discriminative}), which misses a large distribution of negatives (b), our CFCD-Net creating a set of semantically coherent words (concepts) (\eg \{Sofa, table, Chair\} and then learns to discriminate a region from its unrelated concepts. The resulting embedding space better separates the region from its unrelated phrases.}
\label{fig:prior_our_work}
\vspace{-4mm}
\end{figure}

The goal of phrase detection is to localize all instances of a phrase in a database of images.  A key challenge differentiating phrase detection from other phrase grounding tasks is that many (and oftentimes most) of the images do not contain any instances of a phrase. This requires a phrase detector to discriminate between image regions that may contain entities that are semantically similar to a phrase. For example, being able to tell apart ``a semi-truck'' and ``a pickup truck,'' or identifying that a vehicle is ``a gray car'' and not ``a blue car.'' While recent methods have been very successful in the localization-only task (\eg~\cite{Kamath_2021_ICCV,li2021grounded}), phrase detection remains a significant challenge.  This is due, in part, to the long-tailed distribution of phrases, \ie, there may be tens of thousands of annotated queries in a dataset, but many occur only a few times.  %Thus, learning to identify when a phrase is not relevant to an image is key to good performance. 
Prior work used hard-negative sampling to boost performance (\eg,~\cite{hinami-satoh-2018-discriminative}).  However, this generates many false negatives during training, and, as illustrated in Figure~\ref{fig:prior_our_work}(a), the larger pool of semantically similar negative phrases may still embed near the ground truth phrase, causing many false positives during inference.

%this does not align well with downstream tasks such as image captioning, which requires reasoning about fine-grained differences (\eg yacht vs. boat) that may not be addressed by a limited subset of possible negative phrases. Therefore, we address the more relaxed and realistic variant: phrase detection~\citep{Plummer_2020}, where each phrase gets compared to every other phrase in the dataset. 
\begin{figure*}[t]
\centering
\includegraphics[width=0.8\linewidth]{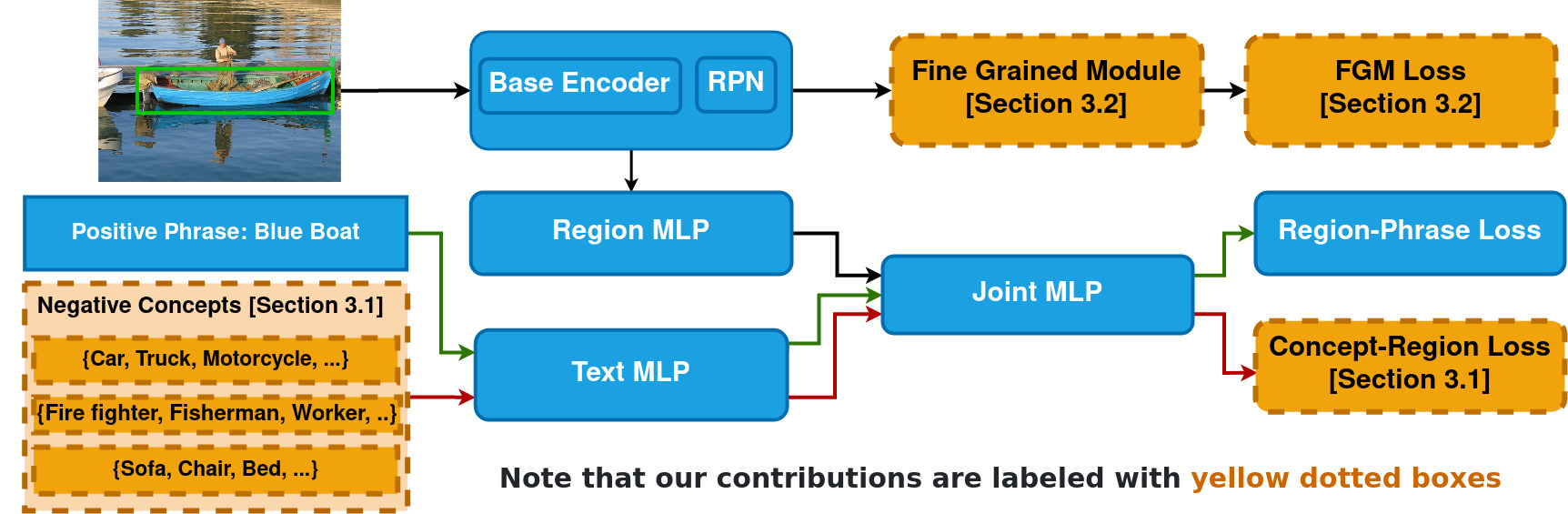}
\vspace{-2mm}
\caption{\textbf{CFCD-Net overview}. We begin by passing the input image to a base encoder (ResNet-101 \cite{he2015deep}) and a region proposal network~\cite{ren2016faster} to generate a set of candidate boxes. To learn a more discerning region-phrase model, we introduce an approach that trains a model to discriminate between a set of automatically constructed visually-related coarse concepts (Section \ref{sec:concept_gen_assi}). We also introduce a fine-grained reasoning module (Section \ref{sec:fgm}) that forces the CFCD-Net to delineate between mutually-exclusive fine-grained tokens.}
\vspace{-2mm}
\label{fig:model}
\end{figure*}

To address this issue, we introduce a Coarse-to-Fine-grained Concept-based Discrimination Network (CFCD-Net), which uses semantically coherent groupings of words (concepts) to perform both coarse and fine grained discrimination. A straightforward approach to distinguish between a large set of phrases would be simply using all the phrases in a batch during training. However, this approach has two major drawbacks. First, phrase grounding datasets are very sparsely labeled, so many unannotated phrases can simply be false negatives (see supplementary for discussion). Second, using all phrases in a single batch would not fit into GPU memory as phrase grounding datasets can have tens or hundreds of thousands of unique training phrases~\cite{Plummer_2020}.  

Prior work avoids memory issues by augmenting a batch with a limited number of positive phrases~\cite{Plummer_2020} or negative phrases~\cite{hinami-satoh-2018-discriminative} during training. However, as mentioned earlier, this may result in semantically similar phrases being embedded nearby the ground truth phrases, reducing performance. Instead, CFCD-Net separates the words belonging to the training phrases into groups with the same part-of-speech.  Then we automatically construct a set of concepts, \ie, semantically coherent bags of words, for each group. For the concepts related to nouns (\eg, (dog, cat, horse) and (car, truck, bike)), we perform a coarse discrimination. % %, as words within the same concept are more likely to be false negatives, but 
%where phrases with words belonging to a different noun concept are used as negatives for each other. More concretely, 
First, for each phrase we obtain a set of unrelated concepts (\ie, concepts that share no semantically similar nouns with the phrase), which we refer to as negative coarse concepts (NCC). Then, given a region/phrase pair, we use the phrase's NCCs as negatives. As illustrated in Figure \ref{fig:prior_our_work}(b), this encourages separation between the region and a wide array of unrelated phrases. 

This approach enables CFCD-Net to minimize false negatives during training, as most unannotated positive phrases belong to the same concept (\eg, the phrase ``a cow'' would not use the concept ``domestic animals'' as a negative). In addition, since the set of concepts is small ($<70$ in our experiments), we can include all of them in every batch. Our approach is similar in spirit to the distributional sampling approach of Wu~\etal~\cite{wu2018sampling}, which demonstrated that balanced batches representing the entire dataset boosts performance over hard-negative mining. However, we create concept groups only once in a short preprocessing step (only a few seconds on a CPU). In contrast, Wu~\etal periodically computes pairwise distances between all instances in the training set, which would take hours on a single GPU every time it was recomputed (details in Section~\ref{sec:results_discussion}).

While NCCs provides a more robust representations of nouns (which typically refer to objects), another challenge in phrase detection the need to recognize attributes, which are commonly captured by adjectives in phrases. We find that adjective concepts work in the opposite way as nouns, where within an adjective concept the words refer to difficult fine-grained differences, while phrases can have words in multiple concept bags.  For example, let us assume we have concepts for colors (\eg, (red, green)) and some texture patterns (\eg, (striped, plaid)).  A shirt could be both red and striped, but a striped shirt should not also be a plaid shirt. We take advantage of these adjective concepts containing mutually exclusive words by adding a fine-grained module (FGM) to our model. Instead of discriminating between the concepts themselves as outlined by our method NCC above, FGM differentiates between the fine-grained words within the concept and then augments the main model score with its prediction.  While there are other potentially useful parts-of-speech that may belong to a phrase (\eg, verbs), we find that existing datasets contain too few of them to make a significant impact on performance.  Figure~\ref{fig:model} provides an overview of our approach.

% A critical step for the success of our method is the correct assignment of phrases to their unrelated concepts. A simple approach would be to use K-Means clustering over a language embedding like BERT~\citep{devlin2019bert} or GLoVE~\citep{pennington-etal-2014-glove}, and then compute similarity between the phrase and a concept. However, as our experiments show, this results in noisy assignment which then reflected on poor detection performance. This is likely because these language embeddings were trained to embed words used in similar contexts near each other but not to distinguish between visually different nouns (\eg they embed \emph{school} and \emph{teacher} nearby since they are related, even though they are not visually similar). In addition, K-means forces all nouns to belong to a cluster, even the outliers, which we found often results in incoherent groups. To avoid these issues, we use visually discriminative language features to represent words. Then we use density-based clustering to create a set of visually coherent concepts, which we assign to phrases based on their semantic similarity.

Our contributions can be summarized as: 
%\vspace{-2mm}
\begin{itemize}[nosep,leftmargin=*]
    \item A novel model, CFCD-Net, that improves over SOTA by 1.5-2 average mAP on phrase detection by mining semantically coherent concepts and then learning to discriminate between a region of interest and its unrelated negative coarse concepts (NCC). 
    \item A fine grained reasoning module (FGM) that boosts performance by 1-4 average mAP over affected phrases by learning to distinguish visually similar instances.
    \item A novel method for automatically mining semantically coherent concepts that are visually similar and with minimum outliers that improves the distribution of our minibatches and, thus, better represents the training data. 
    % \item A thorough discussion of different phrase grounding methods and potential downstream applications.
\end{itemize}

\begin{figure*}[t]
\centering
\includegraphics[width=0.88\linewidth]{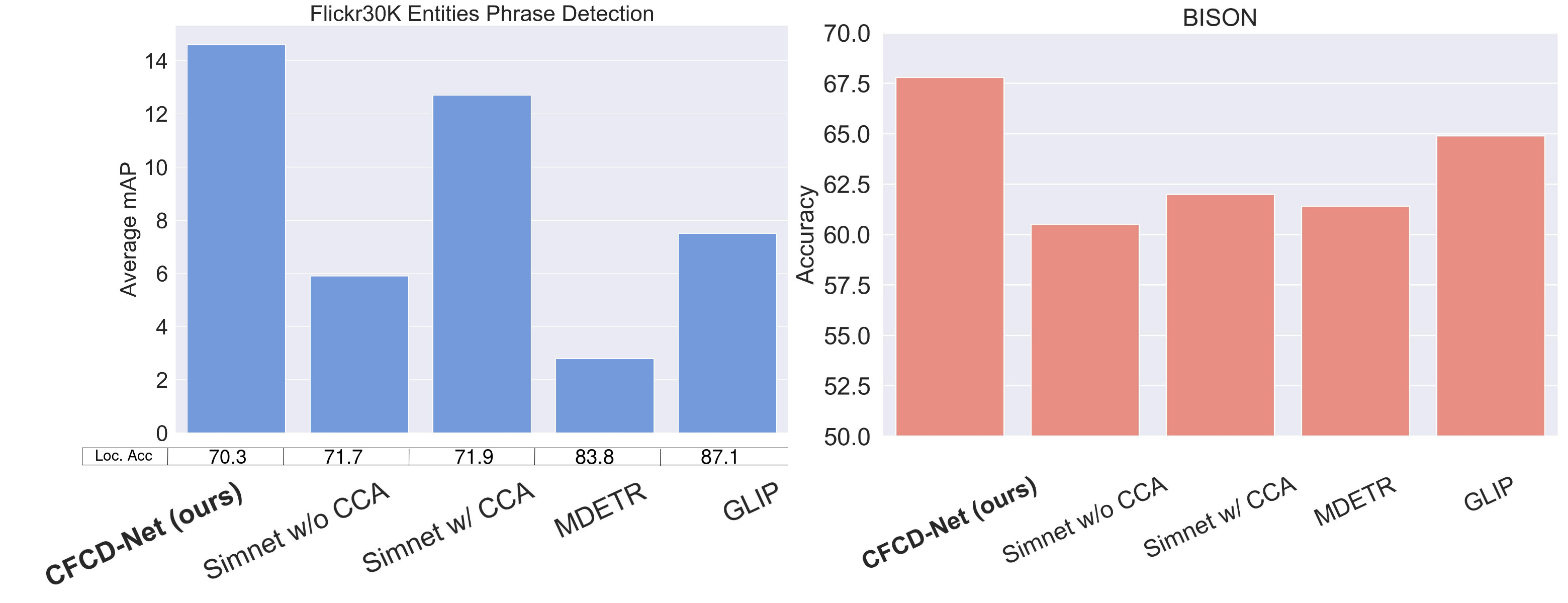}
\vspace{-2mm}
\caption{Comparison across methods~\cite{Plummer_2020,Kamath_2021_ICCV,li2021grounded} on phrase grounding (left) and Binary Image Selection (BISON)~\cite{Hu2019BinaryIS} (right).  Methods along the x-axis are ordered by their localization-only performance. We see that methods designed for localization are not correlated with detection performance (left).  However, we see comparing the left and right plots that improvements on phrase detection are correlated with improved performance on downstream tasks like BISON.  See Section~\ref{sec:related} for discussion}
\vspace{-4mm}
\label{fig:detection_motivation}
\end{figure*}

%% file: 3_related_work.tex
\section{Related work}% and discussion on applications}
\label{sec:related}

%Most prior work addresses phrase localization, where each phrase the goal is to locate a phrase within a ground truth image \citep{10.1145/3078971.3078976,8237357,fukui-etal-2016-multimodal,7780862,7780378,kazemzadeh-etal-2014-referitgame,10.1007/978-3-030-01258-8_16,plummer2017phrase,7410660,wang2018learning,10.1007/978-3-319-46484-8_42,yeh2018interpretable,Wang_2019_ICCV,Kamath_2021_ICCV,li2021grounded}, or cases where a limited set of negative phrases (\eg, only the phrases) or negative images were considered \citep{hinami-satoh-2018-discriminative,zhang2017discriminative,Kamath_2021_ICCV,li2021grounded}. This restrictive evaluation does not align well with how these methods might be used for downstream tasks.  For example, image captioning requires reasoning about whether an image contains ``a yacht'' vs ``a small boat.'' However, methods from prior work were only evaluated on if they could localize the boat, but not if they could determine which phrase is relevant to the image.

In this paper, we refer to phrase grounding as the general task of relating phrases to image regions.  The most popular phrase grounding tasks are phrase localization~\cite{7410660} and referring expression comprehension~\cite{kazemzadeh-etal-2014-referitgame,10.1007/978-3-319-46475-6_5,7780378}.  In these tasks, methods given a ground truth image-phrase pair and must identify the relevant image region (\eg,~\cite{Bajaj_2019_ICCV,hu2015natural,8237357,kazemzadeh-etal-2014-referitgame,plummer2017phrase,10.1007/978-3-030-01258-8_16,yang2019fast}).  This localization-only evaluation does not align well with how these methods might be used for downstream tasks.  For example, image captioning requires reasoning about whether an image contains ``a yacht'' vs ``a small boat.'' However, these methods are typically only evaluated on if they could localize the boat, but not if they could determine which phrase is relevant to the image.  

In contrast, phrase detection methods are ranked precisely on their ability to determine relevance as well as localize the phrase.  Plummer~\etal~\cite{Plummer_2020} demonstrated that using hard-negative mining to boost phrase detection performance was challenge due to the high number of false negatives due to the sparse annotations in phrase grounding datasets.  Instead, they proposed an approach that filled alleviated some label sparsity issues by augmenting the datasets with likely unannotated positive phrases.  In contrast, our CFCD-Net reduces the number of false negatives suffered from hard-negative mining by grouping together semantically similar words (\eg, ``dog, cat, horse'') so that any likely false negatives would be within the same concept.  Thus, any concepts not related to a phrase could safely be used as negatives during training, boosting performance.
\smallskip

\noindent\textbf{Generalizing across phrase grounding tasks.} Plummer~\etal~\cite{Plummer_2020} demonstrated that many methods designed for the localization-only are prune to overfitting to that setting, \ie, they improve performance on localization, but reduce phrase detection performance.  This is partly because distinguishing between similar phrases in the localization-only task is unnecessary since most images only contain a reference to a single object of the same type~\cite{Plummer_2020}.  Thus, a localization model that is given the phrase \emph{a young teenager} as input could look for the object category ``person'' to identify the right object most of the time, whereas a detector also has to determine if it exists at all.  As shown in Figure~\ref{fig:detection_motivation}(left), this means that improved localization performance does not guarantee improvements on detection.  

There is also some work that falls between phrase detection and localization-only tasks, where a ground truth image-phrase pair is not provided~\cite{hinami-satoh-2018-discriminative,Fang_2019_CVPR,zhang2017discriminative}, but they severely limit the number of negative phrases seen by each image at test time.  Thus, methods from these tasks often do not generalize to phrase detection~\cite{Plummer_2020}.
\smallskip

\noindent\textbf{Effect on downstream applications.} We argue the disconnect between localization-only and phrase detection performance partially explains why improving localization performance leads to small improvements on downstream tasks~\cite{liu2016attention,Datta_2019_ICCV,7410660}. Once any grounding method is integrated, even doubling localization performance leads to negligible differences downstream performance~\cite{7410660}. % like Binary Image Selection (BISON) ~\citep{Hu2019BinaryIS}.  
Thus, most vision-language work uses object and attribute detectors to represent images, such as the commonly used bottom-up features~\cite{anderson2018bottomup}, rather than models trained for localization.  In contrast, since phrase detection can also identify what phrases are relevant to an image, it is directly correlated with improved downstream task performance. To illustrate this, we compared several grounding methods on the Binary Image Selection (BISON)~\cite{Hu2019BinaryIS} benchmark. In this task a model is given an sentence and is asked to to choose between two semantically similar images. In our experiment, we extracted the set of noun phrases from the given sentence, and then ranked images by averaging the grounding scores of the phrases. In Figure~\ref{fig:detection_motivation}(right), we see methods that performed well on phrase detection also did well on BISON. See the supplementary for additional details.%Refer to Appendix \ref{sec:image-sentence-supp} for more details.  
\smallskip

\noindent\textbf{Comparison to open-vocabulary detection.} Some recent work (\eg,~\cite{Kamath_2021_ICCV,li2021grounded}) also evaluated on long-tailed object detection benchmarks like LVIS \cite{gupta2019lvis}. However, phrase detection is considerably more challenging, as phrases include information about attributes (\eg, ``a \textbf{red} shirt'') and spatial relationships (\eg, ``a cup \textbf{on top of} a white table''), in addition to a long tailed set of object categories.  Thus, as our experiments show, methods that work well on LVIS (\eg,~\cite{li2021grounded}) often do not generalize to phrase detection.

%% file: 4_method.tex
\section{Coarse-to-Fine-grained Concept-based Discrimination Network (CFCD-Net)}
\label{sec:method}

The goal of phrase detection is to detect and localize all instances of a phrase within a dataset of images. More formally, assume we have a dataset of images $X$ where for each $x \in X$, regions of interests $R_x$ are annotated with boxes $B_x$ and phrases $P_x$. Denote the space of possible phrases as $P = \bigcup\limits_{x\in X} P_x $. Thus, for each $p \in P$ and $x \in X$, the task of phrase detection involves determining whether $p$ is relevant to $x$, and if so localize it with a bounding box. Note that this task involves zero-shot evaluation by definition, since while there are phrases shared between $P_{train}$ and $P_{test}$, $P_{train} \neq P_{test}$, as well as evaluating few-shot and common phrases. Moreover, as we pointed out earlier, detection is a generalization of localization. More concretely, in localization, for each image $x$, only $P_x \subset P$ are evaluated, whereas, in detection, the entirety of $P$ is used. This difference is key for the disconnect between improvements on localization and improvements for downstream tasks as we discussed in Section \ref{sec:related}. 

We improve the discriminative power of detection models by using Concepts. We define a concept $c$ as a bag of semantically coherent words that share the same part-of-speech (\eg, nouns). We mine for a set of concepts $C$ where each $c \in C$ represents a unique semantic concept (\eg, Vehicles: \{car, bike, truck\}). Based on these concepts, we introduce two novel methods of discrimination: a coarse method that uses noun based concepts $C^n$ outlined in Section \ref{sec:concept_gen_assi}, and a fine grained methods that uses adjective based Concepts $C^a$  outlined in Section \ref{sec:fgm}. 
 
% \subsection{Concept Definition}
% \label{sec:concept_def}
\subsection{Expanding Batch Coverage through Coarse Negative Concepts (NCC)}
\label{sec:concept_gen_assi}

The goal of phrase detection is ensuring a model can discriminate between a region of interest $r$ and all unrelated phrases in $P$. Thus, a natural solution to this problem is to expand the batch coverage to include all possible phrases $P$ as potential negatives. However, there can be tens or hundreds of thousands of unique phrases, making it impossible to fit all of them in a single due to GPU memory constraints. Hinami and Satoh~\cite{hinami-satoh-2018-discriminative} sampled phrases using hard-negative mining to increase their model's discriminative power. However, as we discussed in the introduction, these methods miss a large pool of negative phrases that are not "hard" but may still end up embedded reasonably close to the ground truth region embedding.  In addition, phrase grounding datasets are sparsely labeled. Thus, many selected hard negatives may be unannotated positive phrases, \eg, a region labeled as \emph{skier} but not \emph{woman}. %Moreover, since phrase grounding datasets contain tens or hundreds of thousands of phrases, we can not fit all of them in one batch due to GPU memory constraints. 

To mitigate this problem, we propose grouping the nouns in the training dataset into semantically coherent groups, \ie noun based concepts $C^n$. Then, we pair every phrase in the dataset with its set of unrelated concepts. Consequently, given a positive phrase-region pair $(p,r)$, most of the unannoated positive phrases will be contained in one concept (\eg \emph{domestic animals}). Thus, we are more confident that all unrelated concepts should be true negatives. Furthermore, since our concepts group only nouns, we are able to limit the concepts to at most 70 per dataset. This makes it feasible to fit all the concepts in one batch. Moreover, since our concepts span the majority of the language space in the training set, our batch now has a balance of easy to hard negatives, which has been shown to improve performance in prior work \cite{wu2018sampling}. We outline our full pipeline of concept generation and phrase-concept assignment below. 
\smallskip

% To that end, we propose using noun based concepts as defined above and assign every $p \in P$ to its set of unrelated Concepts: $\overline{C_{p_i}}$. Then, For each ground truth phrase/region pair $(p_i, r_i)$ in a given batch, we use the concept set that the phrase is unrelated to $\overline{C_{p_i}} \subset C$ to obtain negative concept/region pair $(\overline{C_{p_i}}, r_i)$. We outline how we generate these pairs below in Section \ref{sec:concept_gen_assi}

% By grouping semantically coherent words, we minimize the chance that a concept is a false negative with respect to an image region. 

% Moreover, a negative concept exposes the region to a broader part of the unrelated language distribution than an individual unrelated phrase.
% \subsubsection{}
% \label{sec:concept_gen_assi}

\noindent\textbf{Concept Generation and Assignment:} To obtain the noun based concept set $C^{n}$, we first extract nouns from our dataset and use a language embedding to represent them. Then, we use the embeddings to cluster the nouns into our semantically similar set of concepts. Given the resulting set of concepts $C^{n}$, we assign each phrase to its set of relevant concepts $C^{n}_p$. A simple solution is to pair a phrase with a concept with which it shares a noun. This approach, however, disregards phrase-concept pairs that don't share nouns but are nevertheless viable pairs. Thus, we make these assignments using semantic similarity as detailed below. 

First, given a phrase $p$, and phrase nouns $p_n$, we collect concepts related to phrase $p$ from simple noun-phrase matching: $C^{n}_{p1} = \{c : c \cap p_{n} \neq \emptyset, c \in C^n \}$. We also consider concepts $C^{n}_{p2}$ which share some semantic similarity (\eg \emph{hoody} with concept: \emph{shirt and sweatshirt})). %This method is designed for when a phrase is still relevant to a concept but they do not explicitly share any nouns . 
Specifically, we look up the language representation $\widetilde{n}$ for each noun $n \in p_{n}$. %We will denote this representation as . 
We also compute a concept $c$ representation by averaging its nouns' textual feature vectors. We will denote this as $\widetilde{c}$. Now assume $\textnormal{sim}(\widetilde{n}, \widetilde{c}) = \frac{\widetilde{n} \textnormal{  }\widetilde{c}}{||\widetilde{n}|| \textnormal{  } ||\widetilde{c}||}$, we compute an "association" score as follows: 
\begin{equation}
    \textnormal{Assoc}(n, c) =\small  \frac{\textnormal{exp}(\textnormal{sim}(\widetilde{n}, \widetilde{c})/\tau)}{\sum_{j=0}^{M} \textnormal{exp}(\textnormal{sim}(\widetilde{n}, \widetilde{c})/\tau)}.
    \label{eq:sim_score}
\end{equation}
Thus, the concepts associated with phrase $p$ are: 
\begin{equation}
    C^{n}_{p2} = \{ c : \textnormal{Assoc}(n, c) > \gamma, n \in p_{n}, c \in  C^n\}.
    \label{eq:assign_sim}
\end{equation}
This way, a phrase is assigned to concepts it shares the greatest semantic similarity. Thus, the final set of phrase concepts is: $C^{n}_{p} = C^{n}_{p1} \cup C^{n}_{p2}$.  Note that phrases near multiple concepts will be assigned to all of them. This ensures we are confident in all negative concepts for a phrase.

A traditional choice of clustering-algorithm/textual embedding is GLoVE \cite{pennington-etal-2014-glove}/K-Means. However, this combination would result in many noisy concepts (discussed further in Section~\ref{sec:model_analysis}). Alternatively, we note that textual embeddings that are visually grounded and designed to make fine-grained distinctions, like ViCO \cite{gupta2019vico}. In addition, density based clustering, like DBSCAN \cite{9356727}, only clusters items that are very close together rather than forcing every item to belong to a cluster like K-means, so we can be more certain they belong to the same concept. As we will show, these choices enabled us to produce a set of concepts with minimal noise. This, in turn, has improved our concept-phrase assignment process in Equation \ref{eq:assign_sim}. In addition to ViCO, we experimented with transformer based embeddings such as BERT \cite{su2019vlbert}. However, we noticed that there was no semantic similarity in the generated concepts. This is because individual nouns do not have enough context which is critical for transformers to work well. Therefore, we do not report the performance of these embeddings in our experiments. 

After assigning each phrase $p$ to its set of related concepts $C_p$, we use the phrase's unrelated concepts $\overline{C_p} = C \backslash C_p$ as negative samples. More concretely, we pair $\overline{C_p}$ with the region $r$ associated with phrase $p$ to obtain negative concept-region pairs. These pairs are concatenated with the ground truth phrase-region positive pairs during training.

% \subsubsection{Expanding Batch Coverage through Concepts}
% \label{sec:expanding_batch}

% Let $s^{c}$ be region-negative concept score, $L$ be the total number of negative concept-region pair, then the loss for the negative coarse concepts (NCC) is:

% \begin{equation}
%     \label{eqn:cr}
%     L_{NCC} = \sum_{i=1}^{L} \log(1 + \exp(s^{c}_i))
% \end{equation}

\subsection{Concept based Fine-Grained Discrimination}
\label{sec:fgm}

The mined negative concepts obtained in Section \ref{sec:concept_gen_assi} which improve our CFCD-Net's discriminative power against a wide spectrum of nouns, but phrases may contain other important cues, such as attributes found in adjectives. However, noun concepts refer to semantically similar objects (\ie, potential unannotated positives), whereas adjective concepts group together words that refer to fine grained differences (\eg, color, texture patterns). Thus, we can not discriminate between concepts like we did for nouns since a phrase can have multiple adjectives from multiple concepts (\eg, \emph{red striped shirt}). However, we can confidently differentiate words within a single concept since its members are often mutually exclusive (\eg, red vs.\ green). We obtain the set of adjective concepts $C^{a}$ by replacing nouns with adjectives in the procedure outlined in Section \ref{sec:concept_gen_assi}.  

Having obtained the set of adjective concepts, we introduce a novel module, Fine Grained Module (FGM), to discriminate between each concept members. Formally, the FGM module encodes image regions using a set of convolutional layers. It then performs multi-label classification on the members of each adjective based concept (\eg, colors) $c \in C^{a}$, then uses them to augment the main model region-phrase scores. Formally, given concept $c \in C^{a}$, $a \in c$, let $R$ be the number of regions/fine-grained adjective pairs, $s^a$ be the region-adjective score, and $l^a$ be its 0/1 label indicating whether it is a positive/negative region-adjective pair, then: 
\begin{equation}
    \label{eq:fgm}
    L_{FGM} = \sum_{i}^{R} l^{a}_{i} \log{s^{a}_{i}} + (1 - l^{a}_{i})\log(1-s^{a}_{i})).
\end{equation}

% \noindent With this, the final loss for CFCD-net is: 
% \begin{equation}
%     \label{eq:final_loss}
%     L_{final} = L_{base} + \lambda_{NCC} L_{NCC} + L_{FGM}
% \end{equation}

\begin{table*}[t]%{0.8\linewidth}
        \setlength{\tabcolsep}{3pt}
        \centering
        \begin{tabular}{rlcccccccc}
        \toprule
         &  & \multicolumn{4}{c}{Flickr30K Entities \cite{7410660}} & \multicolumn{4}{c}{RefCOCO+ \cite{10.1007/978-3-319-46475-6_5}} \\ %\midrule
         %\cmidrule{3-10}
         \midrule
         &  & zero-shot & few-shot & common &  & zero-shot & few-shot & common &  \\
         & \#Train Samples & 0 & 1-100 & $>100$ & mean& 0 & 1-100 & $>100$ & mean \\ \midrule
        \textbf{(a)} & QA R-CNN~\cite{hinami-satoh-2018-discriminative} & 3.9 & 4.3 & \multicolumn{1}{c}{8.9} & 5.7 & 0.9 & 1.5 & \multicolumn{1}{c}{9.3} & 3.9 \\
         & Subquery~\cite{yang2020improving} & -- & -- & \multicolumn{1}{c}{--} & -- & 0.7 & 1.3 & \multicolumn{1}{c}{9.2} & 3.7 \\
         & FAOG~\cite{Wang_2019_ICCV} & 3.2 & 3.5 & \multicolumn{1}{c}{7.6} & 4.8 & 0.7 & 1.1 & \multicolumn{1}{c}{8.9} & 3.6 \\
        \multicolumn{1}{l}{} & MDETR~\cite{Kamath_2021_ICCV} & 1.5 & 2.1 & \multicolumn{1}{c}{4.8} & 2.8 & 1.4 & 2.6 & \multicolumn{1}{c}{10.1} & 4.7 \\
        \multicolumn{1}{l}{} & R-CLIP~\cite{cai2022xdetr} & 8.8 & 6.2 & \multicolumn{1}{c}{3.9} & 6.3 & 4.2 & 3.5 & \multicolumn{1}{c}{2.8} & 3.5 \\
        \multicolumn{1}{l}{} & GLIP~\cite{li2021grounded} & 4.3 & 6.9 & \multicolumn{1}{c}{11.3} & 7.5 & 2.0 & 4.5 & \multicolumn{1}{c}{13.0} & 6.5 \\
        \multicolumn{1}{l}{} & SimNet~\cite{wang2018learning} & 4.7 & 4.4 & \multicolumn{1}{c}{8.6} & 5.9 & 2.0 & 3.3 & \multicolumn{1}{c}{13.1} & 6.1 \\
        \multicolumn{1}{l}{} & CCA~\cite{7410660} & 8.6 & 10.5 & \multicolumn{1}{c}{17.2} & 12.1 & 5.7 & 8.4 & \multicolumn{1}{c}{20.3} & 11.5 \\ \midrule
        \textbf{(b)} & SimNet w/CCA~\cite{Plummer_2020} & 9.7 & 11.2 & \multicolumn{1}{c}{17.3} & 12.7 & 6.0 & 10.2 & \multicolumn{1}{c}{20.1} & 12.1 \\
        \multicolumn{1}{l}{} & + NCC & 10.2 & 11.9 & \multicolumn{1}{c}{18.1} & 13.5 & 6.1 & 10.2 & \multicolumn{1}{c}{21.7} & 12.7 \\
         & + NPA & 10.1 & 12.0 & \multicolumn{1}{c}{18.9} & 13.7 & 5.9 & 10.5 & \multicolumn{1}{c}{22.9} & 13.1 \\
        \multicolumn{1}{l}{} & + NPA + NCC & 10.6 & 12.6 & \multicolumn{1}{c}{19.6} & 14.1 & 6.3 & 10.5 & \multicolumn{1}{c}{23.4} & 13.4 \\
         & CFCD-Net (ours, NPA + NCC + FGM) & \textbf{10.9} & \textbf{12.9} & \multicolumn{1}{c}{\textbf{19.8}} & \textbf{14.6} & \textbf{6.6} & \textbf{10.6} & \multicolumn{1}{c}{\textbf{23.7}} & \textbf{13.7} \\ \midrule
         & \# Unique Phrases & 1783 & 2764 & \multicolumn{1}{c}{472} & 5019 & 5653 & 2293 & \multicolumn{1}{c}{48} & 7994 \\
         & \# Instances & 1860 & 4373 & \multicolumn{1}{c}{8248} & 14481 & 5758 & 3686 & \multicolumn{1}{c}{1171} & 10615 \\ \bottomrule
        \end{tabular}
        \vspace{-2mm}
    \caption{mAp Split by frequency of training instances. \textbf{(a)} contains results reported in prior work or produced using their code.  \textbf{(b)} contains ablations of our model that compares the performance of our its three  components (NPA, NCC, and FGM). See Section \ref{sec:results_discussion} for discussion}
    \label{tb:main_results}
    \vspace{-2mm}
\end{table*}

% \subsubsection{Model inference with FGM}

\noindent\textbf{Module Inference with FGM:} At test time, the FGM module's scores are augmented with the base model's phrase-region scores. Given a phrase $p$, the phrase adjectives $p_a$ and $a\in p_a$, a phrase-region pair score $s^p$, and adjective-region score $s^{a}$, then the final score $s^f$:
\begin{equation}
    s^{f} = (1-\lambda_{c})s^{p} + \lambda_{c}s^{a},
\label{eq:final_score}
\end{equation}
\noindent where $\lambda_{c}$ is a scalar that applies for every $a \in c$. With this, the final loss for CFCD-net is: 
\begin{equation}
    \label{eq:final_loss}
    L_{final} = L_{base} + L_{FGM},
\end{equation}
where $L_{base}$ is the loss used to train the phrase detection backbone which our model is agnostic to.

\subsection{Hard Negative Mining}
Using our proposed Negative Coarse Concepts (NCC) we ensure we learn to discriminate between representatives of the full distribution of possible phrases.  However, discriminating between closely related phrases (\eg car vs.\ truck) is important for good performance.  Despite the noise hard-negative mining methods have shown to produce in phrase grounding~\cite{Plummer_2020}, the benefits have been shown to overcome these issues~\cite{hinami-satoh-2018-discriminative}.  Thus, we adapt Negative Phrase Augmentation (NPA) to help our model distinguish between phrases that are often confused with each other. For each phrase $p$ in the validation set, we record the non-ground truth regions that the model is likely to associate $p$ with. Then, we register the phrases associated with these regions as hard-negative candidates for $p$. We store these candidates in a "Confusion Table" and update it every 3 epochs. Following \cite{hinami-satoh-2018-discriminative}, we reduce some noise from known mutually non-exclusive phrases using WordNet \cite{10.1145/219717.219748}.  Specifically, we remove any potential hard-negatives that share words with a parent-child relationship in the WordNet hierarchy.  In other words, the phrase \emph{vehicle} would remove \emph{car} as a potential hard-negative since they hold parent-child relationship.  After this process, for each positive phrase-region pair $(p,r)$ in the batch, we sample a candidate hard negative phrase $\overline{p}$ and concatenate the negative pair $(\overline{p},r)$ to the batch.

%% file: 5_experiments.tex
\begin{table}[t]
    \centering
    \setlength{\tabcolsep}{1pt}
    \begin{tabular}{llc}
    \toprule
    Model & Training Data & Inference Time \\ \midrule
    MDETR \cite{Kamath_2021_ICCV} & COCO, VG, F30K Entities(200k) & 7 Days \\ 
    GLIP  \cite{li2021grounded} & FourODs,GoldG+, COCO(24M) & 1 Day \\
    ours & F30K Entities(30k) & $<1$ hour \\ \bottomrule
    \end{tabular}
    \vspace{-2mm}
    \caption{Comparison between CFCD-Net and transformer based phrase localization models evaluated on Flickr30K Entities. See Section \ref{sec:results_discussion} for discussion}
    \vspace{-2mm}
    \label{tb:model_details}
\end{table}

\begin{table*}[t]%[h!]
    \setlength{\tabcolsep}{3pt}
    \centering
    \begin{tabular}{lcccccccc}
    \toprule
     & \multicolumn{4}{c}{Flickr30K Entities \cite{7410660}} & \multicolumn{4}{c}{RefCOCO+ \cite{10.1007/978-3-319-46475-6_5}} \\ %\hline
     \midrule
    & zero-shot & few-shot & common &  & zero-shot & few-shot & common &  \\
    \#Train Samples & 0 & 1-100 & $>100$ & mean& 0 & 1-100 & $>100$ & mean \\ \midrule
    %\midrule
    SimNet w/CCA \cite{Plummer_2020} & 13.0 & 14.7 & \multicolumn{1}{c}{12.0} & 13.2 & 7.0 & 10.3 & \multicolumn{1}{c}{11.0} & 9.4 \\
    + NPA + NCC & 12.9 & 13.9 & \multicolumn{1}{c}{17.1} & 14.7 & 7.1 & 10.6 & \multicolumn{1}{c}{17.1} & 11.6 \\
    CFCD-Net (NPA + NCC + FGM) & \textbf{15.0} & \textbf{17.2} & \multicolumn{1}{c}{\textbf{18.5}} & \textbf{16.8} & \textbf{7.8} & \textbf{11.7} & \multicolumn{1}{c}{\textbf{17.9}} & \textbf{12.5} \\ \bottomrule
    \end{tabular}
    \vspace{-2mm}
    \caption{Performance of the phrases impacted by our FGM module compared with the previous SOTA. See Section \ref{sec:results_discussion} for discussion}
    \label{tb:fgm_results}
    \vspace{-2mm}
\end{table*}

\section{Experiments}
\label{sec:experiments}
\noindent\textbf{Datasets:} We evaluate CFCD-net on two common phrase grounding datasets. First, we use Flickr30K Entities \cite{7410660} that consists of 276K bounding boxes in 32K images for the noun phrases associated with each image’s descriptive captions (5 per image) from the Flickr30K dataset \cite{Young2014FromID}. We use the official splits \cite{7410660} that consist of 30K/1K/1K train/test/validation images. Second, we evaluate on RefCOCO+ \cite{10.1007/978-3-319-46475-6_5}, which consists of 19,992 images from the COCO dataset \cite{10.1007/978-3-319-10602-1_48} that have been labeled with 141,564 region descriptions. We use the official split \cite{10.1007/978-3-319-46475-6_5}, which splits the train/val and testing sets 16K/1.5K/1.5K. Both datasets are licensed under creative commons. 
\smallskip

%\begin{figure*}[t!]
%    \centering
%    \includegraphics[width=0.95\linewidth]{images/qualt_full_1.jpeg}
%    \vspace{-2mm}
%    \caption{Comparison between CFCD-Net vs prior work on positive (ground truth) precision scores. See Section \ref{sec:model_analysis} for discussion} 
%    \vspace{-2mm}
%    \label{fig:qualt}
%\end{figure*}

\noindent\textbf{Metrics:} We follow the evaluation protocols of Plummer~\etal~\cite{Plummer_2020}.  For every image we obtain the most likely region and confidence score for every phrase in our test split.  For any ground truth phrases in an image, we consider them successfully localized if the predicted bounding box has at least 0.5 intersection-over-union with its ground truth bounding box.  Then, we compute average precision (AP) for each phrase and then split them into zero-shot, few-shot, and common sets, based on if they didn't occur in our training split, if they had between 1-100 occurrences, or if they occurred more than 100 times, respectively.  We then report an overall mAP for each set of phrases, as well as the average of them for an overall performance score.  This procedure ensures that the zero-shot and few-shot phrases are not over represented compared to the common phrases, since the zero- and few-shot sets have more unique phrases, but represent a smaller portion of overall instances.
\smallskip

\noindent\textbf{Implementation details.} We use the same framework as the state-of-the-art \cite{Plummer_2020} for a fair comparison.  This combines a ResNet-101~\cite{he2015deep} based Faster R-CNN model~\cite{ren2016faster} to obtain a set of image regions.  Phrases are representing using the average of HGLMM Fisher Vectors \cite{hglmm}.  The model is optimized using ADAM \cite{kingma2017adam} and we keep all hyperparameters introduced by prior work fixed.  All hyperparameters introduced by our paper are set via grid search on the validation set (see Section \ref{sec:model_analysis} for a sensitively analysis). This resulted in $\tau = 0.01, \gamma = 0.2$ from Eq.~(\ref{eq:sim_score}) and Eq.~(\ref{eq:assign_sim}), respectively, for both datasets. We set DBSCAN's sensitivity to $0.43$ for Flickr30K and $0.53$ for RefCOCO+. Our model was trained using a single NVIDIA RTX 8000 GPU.
% \smallskip

\subsection{Results}
\label{sec:results_discussion}

Table \ref{tb:main_results} compares our model (CFCD-Net) (averaged over 3 runs) with the state-of-the-art in phrase grounding. Comparing the last line of Table \ref{tb:main_results}(b) to the results from prior work in Table \ref{tb:main_results}(a) and the first line of Table~\ref{tb:main_results}(b) we get 1.5-2 point gain in mAP over the state of the art. This gain came mostly from common phrases, where we achieved a 2-3 point gain over prior work. Furthermore, note the results of transformer-based models MDETR \cite{Kamath_2021_ICCV}, GLIP \cite{li2021grounded}, and an adaptation of CLIP~\cite{CLIP} to phrase grounding, R-CLIP~\cite{cai2022xdetr}.  These methods performed well on phrase localization and object detection benchmarks like LVIS~\cite{gupta2019lvis}. This is further evidence that simply optimizing for the localization objective as in prior work does not yield discriminative models even as data is scaled up. Furthermore, as reported in Table~\ref{tb:model_details}, our model inference time is far shorter.  This is because both MDETR and GLIP use cross modal transformers to fuse language/vision data adding significant computational complexity. In contrast, our computationally efficient fusion based on element-wise product followed by lightweight fully connected layers results in an inference time less than 1.1\% that of either model.  %Notably, GLIP is faster than MDETR. This is because we were able to process more than one phrase at a time by concatenating the phrases together in one sentence (\eg "red car, blue table, \etc") effectively speeding up inference.

In Table \ref{tb:main_results}(b) we also report the contribution of each component of our model. We observe that NCC makes a significant improvements over both datasets. Although the improvements are quite similar to those made by NPA~\cite{hinami-satoh-2018-discriminative}, they come at a fraction of the cost. Specifically, NPA must update a confusion table to select good negatives for every phrase in the training pool during training, which took 3 hours on Flickr and 4 hours on RefCOCO+ using 4 NVIDIA RTX8000 GPUs every time the confusion table was updated (every 3 epochs in our experiments). In contrast, the groups used by NCC are computed only once and takes a few seconds on a CPU using precomputed language features.  Nevertheless, as shown in Table \ref{tb:main_results}(b), we find we get best performance when both methods are combined (NPA + NCC). 

In addition to the gains from using NCC and NPA, Table~\ref{tb:main_results}(b) reports that the FGM module (described in Section \ref{sec:fgm}) further boosts performance for both datasets over all phrases. Note though that the FGM module only affects phrases with adjectives.  Thus, in Table \ref{tb:fgm_results} we report performance over affected phrases, where we report a 3-3.5 gain on Flickr30K Entities and RefCOCO+. See supplementary for qualitative examples of our detection results. %Finally, Figure  \ref{fig:qualt} provides additional insight into our model behavior. For each positive phrase, we compute its precision score using the phrase score as a threshold. Thus, higher precision means that there are fewer false positives. Compared to prior work, our model consistently maintains higher precision for each positive phrase. For example, in Figure \ref{fig:qualt}(b) our model has higher precision for phrase "snowy hillside" in the image on the left than prior work model in Figure  \ref{fig:qualt}(a).

\begin{table*}[t]
    \setlength{\tabcolsep}{3pt}
    \centering
    \begin{tabular}{lcccccccc}
        \toprule
        & \multicolumn{4}{c}{Flickr30K Entities \cite{7410660}}  &  \multicolumn{4}{c}{RefCOCO+ \cite{10.1007/978-3-319-46475-6_5}} \\
        \midrule
        & zero-shot & few-shot & common &  & zero-shot & few-shot & common &  \\
         \#Train Samples & 0 & 1-100 & $>100$ & mean& 0 & 1-100 & $>100$ & mean \\ \midrule
         Baseline [SimNet w/CCA] & 9.7  & 11.2 & 17.3 & 12.7 & 6.0 & 10.2 & 20.1 & 12.1 \\
         \midrule
        NCC w/ K-means + GLoVE & 10.0  & 11.8 & \textbf{18.5} & 13.4 & 6.0 & 9.9 & 20.4 & 12.1 \\
        NCC w/ K-means + ViCo &  10.1 & 11.9 & 18.3 & 13.4 & 6.0 & 9.8 & 20.7 & 12.2 \\
        NCC w/ DBSCAN + GLoVE &  9.9 & \textbf{11.9} & 18.4 & 13.4 & \textbf{6.1} & \textbf{10.4} & 20.7 & 12.4 \\
        NCC w/ DBSCAN + ViCo & \textbf{10.2} & \textbf{11.9}&  18.1 & \textbf{13.5} &  \textbf{6.1} & 10.2 & \textbf{21.7} & \textbf{12.7} \\
        \bottomrule
    \end{tabular}
    \vspace{-2mm}
    \caption{Comparison of the effect of clustering algorithm and language embedding when creating the concepts used by our NCC sampling strategy. See Section~\ref{sec:model_analysis} for discussion.}
     \vspace{-2mm}
    \label{tb:cr_ab}
\end{table*}

\subsection{NCC Hyperparameter Analysis}
\label{sec:model_analysis}

\noindent\textbf{Clustering-Algorithm/Textual-Embedding.} Table \ref{tb:cr_ab} reports the effect of different choices of clustering algorithm and language embedding used to create our NCC concepts (outlined in Section \ref{sec:concept_gen_assi}) has on phrase detection performance.  A traditional choice of clustering-algorithm/textual-embedding is K-Means clustering over GLoVE features~\cite{pennington-etal-2014-glove}, but we show a small but mostly consistent gain with DBSCAN \cite{9356727} + ViCo~\cite{gupta2019vico} instead. 

\begin{figure}[t]
    \centering
    \includegraphics[width=0.95\linewidth]{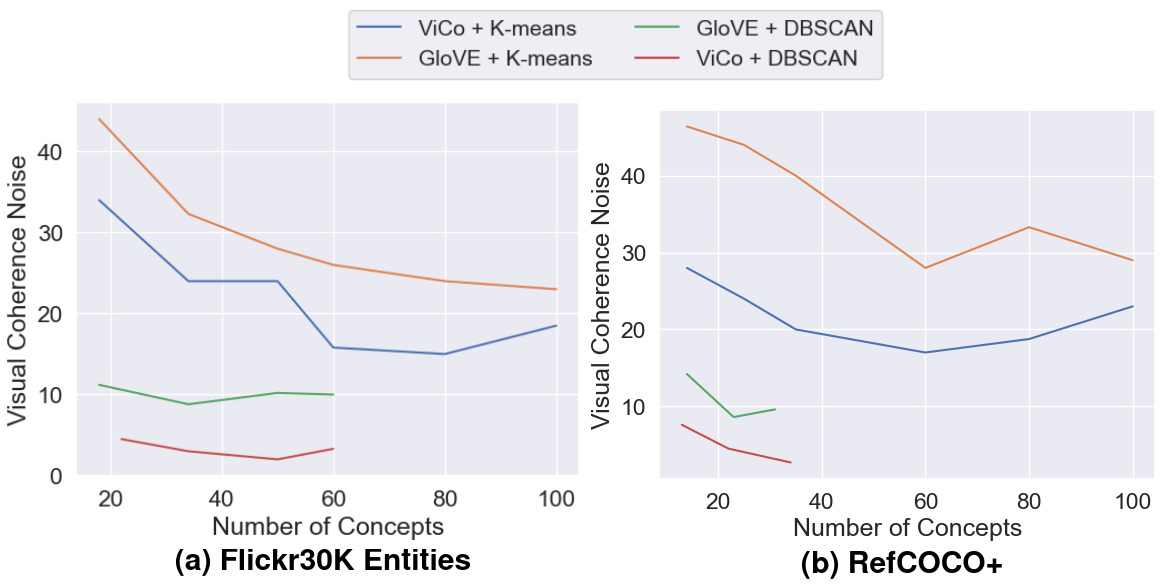}
    \vspace{-2mm}
    \caption{Quantitative comparison between difference choices of clustering-algorithm/textual-embedding for concept generation. Refer to Section \ref{sec:concept_gen_assi} for discussion} 
    \vspace{-2mm}
    \label{fig:visual_noise}
\end{figure}

To better understand the causes in these performance differences, we begin by computing a "visual coherence noise" metric over concepts. We manually inspected each concept and counted the number of times at least \%50 of the words were not visually similar.  Since there are relatively few concepts (less than 100), this only takes a few minutes for a single annotator.  Figure \ref{fig:visual_noise} reports the performance of the different embedding and clustering combinations, where ViCO + DBSCAN also obtains best performance. 

\begin{figure}[t]
    \centering
    \includegraphics[width=0.95\linewidth]{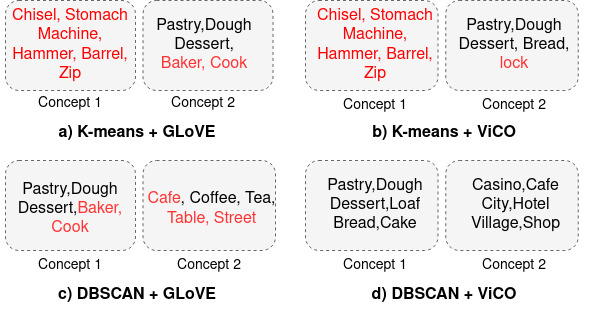}
    \vspace{-2mm}
    \caption{Qualitative comparison between difference choices of clustering-algorithm/textual-embedding for concept generation. Refer to Section \ref{sec:concept_gen_assi} for discussion} 
    \vspace{-2mm}
    \label{fig:visual_noise_qual}
\end{figure}

Figure \ref{fig:visual_noise_qual} provides examples of the the different concepts.  Note that the left concept related to K-means + Glove in Figure \ref{fig:visual_noise_qual}(a) contains largely unrelated words. This is because K-means does not impose a constraint over each cluster density, \ie, outliers in the embedding space are clustered with other concepts even though there is little evidence they belong together. Furthermore, even when the concept is less noisy (concept on the right), it includes words that are semantically similar but not visually similar (\eg Baker vs Dough). This likely harms our concept-phrase assignment process. For example, given a phrase \emph{tall baker}, a textual embedding that is not visually grounded would likely assign a concept containing ("bread, baker, dough") as related to the phrase. However, a visually grounded embedding would only assign concepts that exclusively contain humans as related. ViCO \cite{gupta2019vico} helps to address these shortcomings because it was explicitly trained to consider visual similarity, which should ensure that concepts rule out words that are semantically similar but not visually similar. Furthermore, DBSCAN \cite{9356727}, whose clusters are formed from words that fall within a density threshold $\epsilon$. Thus, the method is more effective at ruling out outliers, as illustrated in Figure \ref{fig:visual_noise_qual}(d), which are more visually similar and include fewer outliers.  %Thus, the performance gain in Table \ref{tb:cr_ab}  %Indeed, ViCO/DBSCAN performs the best, which is then reflected in better performance in Table \ref{tb:cr_ab}. 
\smallskip

\begin{figure}[t]
\centering
\includegraphics[width=0.45\textwidth]{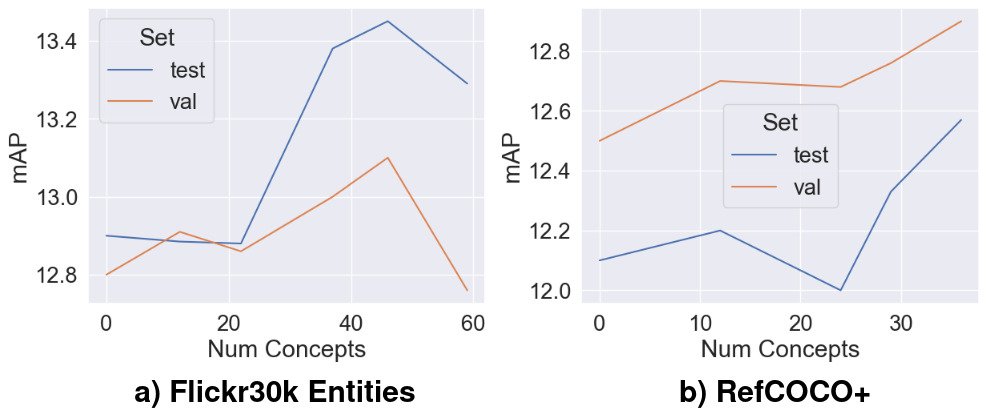}
\caption{Effect of Concept Size $|C|$}
\label{fig:hyper_size}
\vspace{-2mm}
\end{figure}

\noindent\textbf{Number of Concepts.} We also investigate the effect of changing the number of concepts in Figure \ref{fig:hyper_size}. We progressively increase DBSCAN's density threshold for each dataset until we can not generate more concepts. In other words, as we increase the threshold, we create more concepts by force them to be more tightly coupled.  This breaks apart less semantically related words, thus improving the accuracy of our phrase-concept assignment in Section \ref{sec:concept_gen_assi}.
\smallskip

\begin{figure}[t]
\centering
\includegraphics[width=0.45\textwidth]{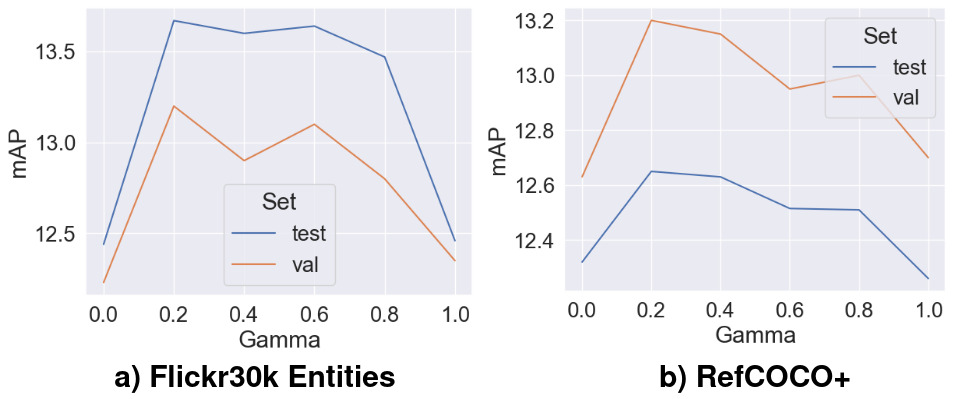}
\caption{Effect of $\gamma$ in Equation \ref{eq:assign_sim}}
\label{fig:hyper_gamma}
\end{figure}

\noindent\textbf{Concept-Phrase Assignment.}  Figure \ref{fig:hyper_gamma} reports the effect of changing $\gamma$ in Eq.~(\ref{eq:assign_sim}), which controls whether a given concept is assigned to a phrase (\ie they are related). We vary $\gamma$ between 0 (a phrase is assigned to every given concept) and 1 (a phrase is only assigned to concepts that it share nouns with). We report best performance for $\gamma$ values between (0.2, 0.6). Performance drops for $\gamma < 0.2$ as we incorrectly assign concepts to many phrases. Performance also drops when $\gamma > 0.8$, demonstrating the importance of our similarity-based matching approach. %\Ie, simply assigning phrases to concepts that only share nouns with is not sufficient. 
As discussed in Section \ref{sec:concept_gen_assi}, this is due to many concepts being relevant to phrases despite not sharing words.


% Third, and finally, we study the effect of changing $\lambda_{EC}$ used in Eq \ref{eq:final_loss} in Figure \ref{fig:hyper_lambda_ec} that controls the contribution of our Concept-Entity loss. Overall, we note that performance for both datasets is best at 0.1.  We note that both the $\gamma$ from Eq \ref{eq:assoc} and $\lambda_{EC}$ from Eq \ref{eq:final_loss} achieved the highest performance at the same parameter value, demonstrating that these hyperparameter val generalize across datasets, whereas CFCD-Net tends to favor larger number of concepts.

%   \begin{minipage}[b]{0.32\textwidth}
%       \begin{subfigure}{ \linewidth}
%         \centering
%         \includegraphics[width=\textwidth]{images/hyper_lambda_flickr.png}
%         \caption{Flickr30K Entities}
%       \end{subfigure}
%       \begin{subfigure}{ \linewidth}
%         \centering
%         \includegraphics[width=\textwidth]{images/hyper_lambda_refcoco.png}
%         \vspace{-2mm}
%         \caption{RefCOCO+}
%       \end{subfigure}
%       \caption{Effect of changing  $\gamma$ in Eq \ref{eq:assoc} has on performance. See Section \ref{sec:model_analysis} for discussion}
%       \label{fig:hyper_lambda_ec}
%   \end{minipage}

%% file: 6_conclusion.tex
\section{Conclusion}

In this work, we introduced a new phrase detection mode (CFCD-NET) that improves performance by 1.5-2 points on two phrase detection datasets.   It does so by incorporating visually coherent clusters (concepts) to sample negative concept-region that effectively improve the model discriminative abilities when compared to prior work. Our model further improves performance by incorporating a novel fine grained module that learns to discriminate between adjective fine grained tokens. Notably, our approach even outperforms recent transformer-based methods like MDETR~\cite{Kamath_2021_ICCV} and GLIP~\cite{li2021grounded} on phrase detection performance while requiring less training data and significantly faster inference speeds. In addition, although our experiments used the Faster R-CNN framework to fairly compare to prior work, our contributions are modular and can be adapted to any underlying detection framework. 
%%%%%%%%%%%%%%%%%%%%%%%%%%%%%%%%%%%%%%%%%%%%%%%%%%%%%%%%%%%%